\documentclass[10pt,twocolumn,letterpaper]{article}

\usepackage{iccv}
\usepackage{times}
\usepackage{epsfig}
\usepackage{graphicx}
\usepackage{amsmath}
\usepackage{amssymb}
\usepackage{url}
\usepackage{soul}
\usepackage{verbatim}
\usepackage{subfigure}
\usepackage{calc}
\usepackage{amstext}
\usepackage{stackengine}
\newcommand\IncG[2][]{\addstackgap{\raisebox{-.5\height}{\includegraphics[#1]{#2}}}}
\usepackage{array}
\usepackage{amsmath,amssymb} 
\usepackage{xcolor}
\usepackage{paralist}
\usepackage[shortlabels]{enumitem}
\usepackage{algorithm}  
\usepackage{algpseudocode}  
\usepackage{amsmath}  
\renewcommand{\algorithmicrequire}{\textbf{Input:}}

\usepackage{comment}

\usepackage[breaklinks=true,bookmarks=false]{hyperref}

\iccvfinalcopy 

\def\iccvPaperID{****} 

\begin{document}

\title{Visualization of Convolutional Neural Networks \\ for Monocular Depth Estimation}

\author{ Junjie Hu$^{1,2}$ \hspace{1cm}  Yan Zhang$^{1,2}$ \hspace{1cm} Takayuki Okatani$^{1,2}$\\
$^1$ Graduate School of Information Sciences, Tohoku University, Japan \\
$^2$ Center for Advanced Intelligence project, RIKEN, Japan\\
{\tt\small \{junjie.hu, zhang, okatani\}@vision.is.tohoku.ac.jp}
}

\maketitle

\begin{abstract}
Recently, convolutional neural networks (CNNs) have shown great success on the task of monocular depth estimation. A fundamental yet unanswered question is: how CNNs can infer depth from a single image. Toward answering this question, we consider visualization of inference of a CNN by identifying relevant pixels of an input image to depth estimation. We formulate it as an optimization problem of identifying the smallest number of image pixels from which the CNN can estimate a depth map with the minimum difference from the estimate from the entire image. To cope with a difficulty with optimization through a deep CNN, we propose to use another network to predict those relevant image pixels in a forward computation. In our experiments, we first show the effectiveness of this approach, and then apply it to different depth estimation networks on indoor and outdoor scene datasets. The results provide several findings that help exploration of the above question.
\end{abstract}

\section{Introduction}
Enabling computers to perceive depth from monocular images has attracted a lot of attention over the past decades. It was shown recently \cite{Eigen2014depth} that employment of deep convolutional neural networks (CNNs) achieves promising performance.
Since then, a number of studies \cite{Li2015DepthAS,Eigen2015PredictingDS,chakrabarti2016depth,chen2016single,laina2016deeper,Xu2017MultiscaleCC,li2017two,fu2018deep,Kendall2017WhatUD} have been published on this approach, leading to significant improvement of  estimation accuracy.

\begin{figure}[t]
\centering
\subfigure {\includegraphics[scale=0.24]{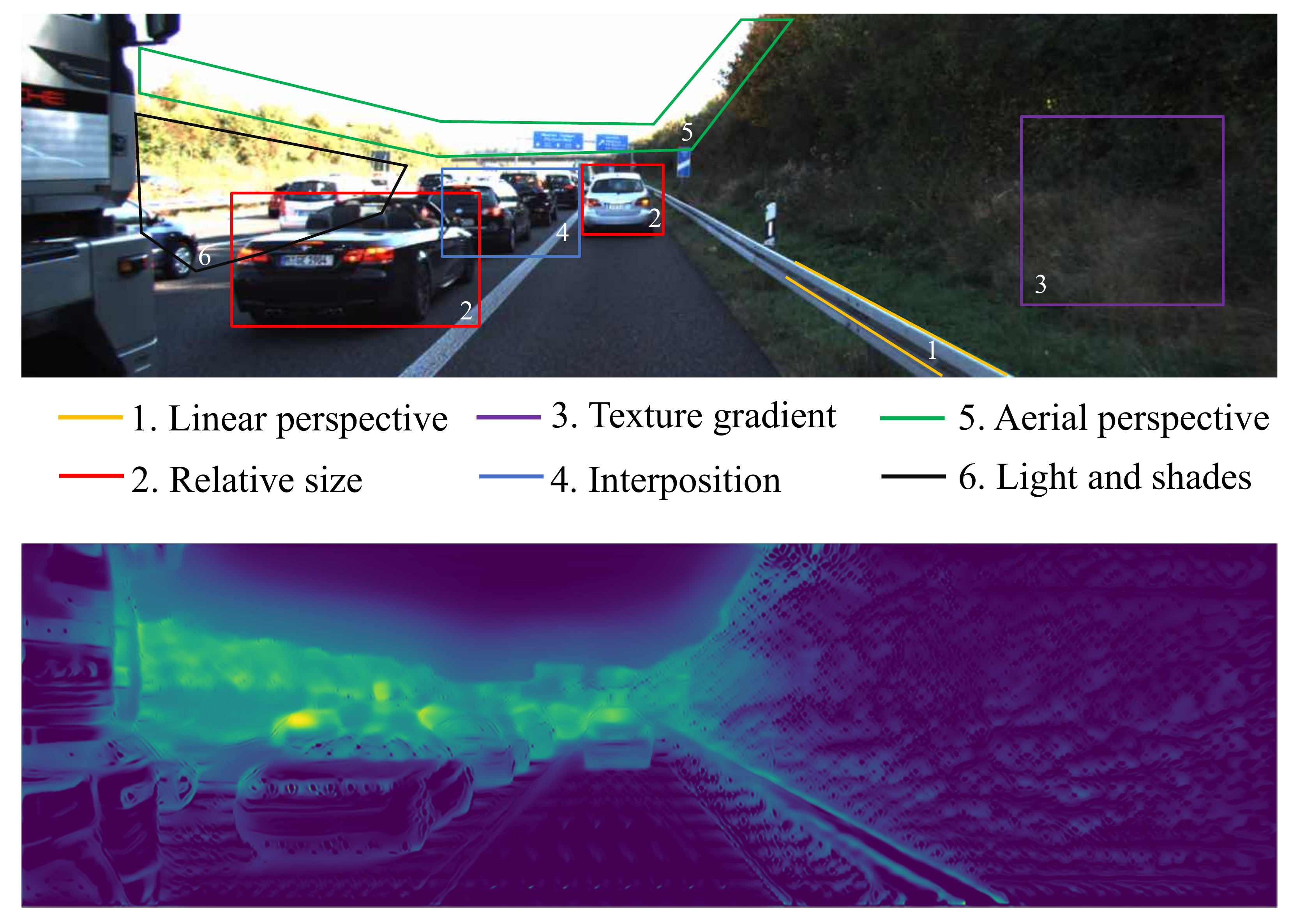}}
\caption{The upper row shows six monocular cues that are considered to be used for depth perception in human vision.  The lower row shows the mask predicted by our method.  
}
\label{fig_cues}
\end{figure}

On the other hand, it is largely unknown why and how CNNs can estimate depth of a scene from its monocular image; they are basically black boxes as in other tasks. 
This will be an obstacle for this method to be employed in real-world applications, such as vision for self-driving cars and service robots, although it could be a cheap alternative solution to existing 3D sensors. 
In these applications, interpretability is essential for safety reasons.

\begin{figure*}[t]
\centering
\subfigure {\includegraphics[scale=0.5]{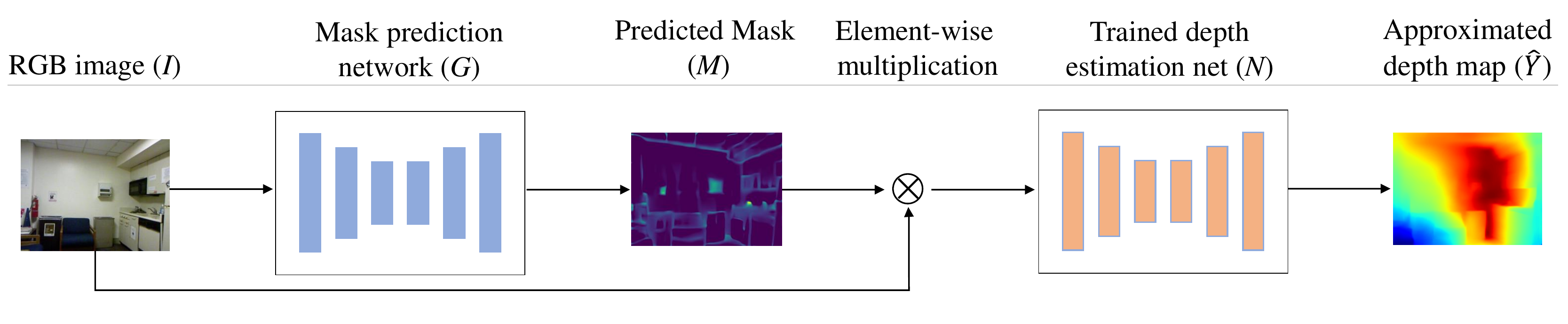}}
\caption{Diagram of the proposed approach. The target of visualization is the trained depth estimation net $N$. To identify the pixels of the input image $I$ that $N$ uses to estimate its depth map $Y$, we input $I$ to the network $G$ for predicting the set of relevant pixels, or the mask $M$. The output $M$ is element-wise multiplied with $I$ and inputted to $N$, yielding an estimate $\hat{Y}$ of the depth map. $G$ is trained so that $\hat{Y}$ will be as close to the original estimate $Y$ from the entire image $I$ and $M$ will be maximally sparse. Note that $N$ is fixed in this process.
}
\label{fig_arch}
\end{figure*}

Long-term studies in psychophysics have revealed that human vision uses several cues for monocular depth estimation, such as linear perspective, relative size, interposition, texture gradient, light and shades, aerial perspective, \etc.
\cite{lebreton2014measuring,howard2002seeing,landy1995measurement,saxena2007depth,reichelt2010depth, kelly1977visual}; an example is given in Figure~\ref{fig_cues}. A natural question arises, 
{\it do CNNs utilize these cues?}
Exploring this question will help 
our understanding of why CNNs can (or cannot) estimate depth from a given scene image. 
To the best of our knowledge, the present study is the first attempt to analyze how CNNs work on the task of monocular depth estimation.

It is, however, hard to find direct answers to the above questions; after all, it is still difficult even with human vision. Thus, as the first step toward this end, we consider visualization of CNNs on the task. To be specific, as in previous studies of visualization of CNNs for object recognition, we attempt to identify the image pixels that are relevant to depth estimation. To do this, we hypothesize that the CNNs can infer depths fairly accurately from only a selected set of image pixels. An underlying idea is an observation with human vision that most of the cues are considered to be associated with small regions in the visual field. 

We then formulate the problem of identifying relevant pixels as a problem of sparse optimization. Specifically, we estimate an image mask that selects the smallest number of pixels from which the target CNN can provide the maximally similar depth map to that it estimates from the original input. This optimization requires optimization of the output of the CNN with respect to its input. As is shown in previous studies of visualization, such optimization through a CNN in its backward direction sometimes yields unexpected results, such as noisy visualization \cite{Simonyan2013DeepInside,Springenberg2014StrivingFS} at best and even phenomenon similar to adversarial examples \cite{Fong2017InterpretableEO} at worst. To avoid this issue, we use an additional CNN to estimate the mask from the input image in the forward computation; this CNN is independent of the target CNN of visualization. Our method is illustrated in Figure \ref{fig_arch}. 

We conduct a number of experiments to evaluate the effectiveness of our approach. We apply our method to CNNs trained on indoor scenes (the NYU-v2 dataset) and those trained on outdoor scenes (the KITTI dataset). We confirm through the experiments that
\begin{itemize}
    \item CNNs can infer the depth map from only a sparse set of pixels in the input image with similar accuracy to those they infer from the entire image; 
    \item The mask selecting the relevant pixels can be predicted stably by a CNN. This CNN is trained to predict masks for a target CNN for depth estimation.
\end{itemize}
The visualization of CNNs on the indoor and outdoor scenes provides several findings including the following, which we think contribute to understanding of how CNNs works on the monocular depth estimation task.
\begin{itemize}
    \item CNNs frequently use some of the edges in input images but not all of them. Their importance depends not necessarily on their edge strengths but more on usefulness for grasping the scene geometry. 
    \item For outdoor scenes, large weights tend to be given to distant regions around the vanishing points in the scene. 
\end{itemize}

\section{Related work}

There are many studies that attempt to interpret inference of CNNs,
most of which have focused on the task of image classification \cite{Cao2015LookAT,Zhou2016LearningDF,Springenberg2014StrivingFS,Smilkov2017SmoothGradRN,Zintgraf2016VisualizingDN,ribeiro2016should,Selvaraju2017GradCAMVE,jetley2018learn,Fong2017InterpretableEO,Mahendran2016SalientDN,Sundararajan2017AxiomaticAF}. 
However, there are only a few methods that have been recognized to be practically useful in the community \cite{Guidotti2018ASO,Kindermans2017TheO,kindermans2017learning}. 

Gradient based methods \cite{Smilkov2017SmoothGradRN,Mahendran2016SalientDN,Sundararajan2017AxiomaticAF} compute a saliency map
that visualizes sensitivity of each pixel of the input image to the final prediction, which is obtained by calculating the derivatives of the output of the model with respect to each image pixel. 

There are many methods that mask part of the input image to see its effects \cite{Zeiler2014VisualizingAU}.  General-purpose methods developed for interpreting inference of machine learning models, such as LIME \cite{ribeiro2016should} and Prediction Difference Analysis \cite{Zintgraf2016VisualizingDN}, may be categorized in this class, when they are applied to CNNs classifying an input image. 

The most dependable method as of now for visualization of CNNs for classification is arguably the class activation map (CAM) \cite{Zhou2016LearningDF}, which calculate the linear combination of the activation of the last convolutional layers in its channel dimension. Its extension, Grad-CAM \cite{Selvaraju2017GradCAMVE}, is also widely used, which integrates the gradient-based method with CAM to enable to use general network architectures that cannot be dealt with by CAM. 

However, the above methods, which are developed mainly for explanation of classification, cannot directly be applied to CNNs performing depth estimation. In the case of depth estimation, the output of CNNs is a two-dimensional map, not a score for a category. This immediately excludes gradient based methods as well as CAM and its variants. The masking methods that employ fixed-shape masks \cite{Zintgraf2016VisualizingDN} or super-pixels obtained using low-level image features \cite{ribeiro2016should} are not fit for our purpose, either, since there is no guarantee that their shapes match well with the depth cues in input images that are utilized by the CNNs.

\section{Method}
\label{method}

\subsection{Problem Formulation} \label{problem_def}

Suppose a network $N$ that predicts the depth map of a scene from its single RGB image as
\begin{equation}
    Y = N(I),
\end{equation}
where 
$Y$ is an estimated depth map and $I$ is the normalized version of the input RGB image. Following previous studies, we normalize each image by the z-score normalization. This model $N$ is the target of visualization.

Human vision is considered to use several cues to infer depth information, most of which are associated with regions with small areas in the visual field. 
Thus, we make an assumption here that {\em CNNs can infer depth map equally well from a selected set of sparse pixels of $I$, as long as they are relevant to depth estimation.}
To be specific, we denote a binary mask selecting pixels of $I$ by $M$ and a masked input by $I \otimes M$, where $\otimes$ denotes element-wise multiplication. The depth estimate $\hat{Y}$ provided by our network $N$ for the masked input  is
\begin{equation}
    \hat{Y} = N(I \otimes M).
\end{equation}
Our assumption is that $\hat{Y}$ can become very close to the original estimate $Y=N(I)$, when the mask $M$ is chosen properly.

Now, we wish to find such a mask $M$ for a given input $I$ that $\hat{Y}=N(I\otimes M)$ will be as close to $Y=N(I)$ as possible. As our purpose is to understand depth estimation, we also want $M$ that is as sparse  as possible (\ie, having the smallest number of non-zero pixels). To do so, we relax the condition that $M$ is binary, \ie, its element is either 0 or 1. We instead assume each element of $M$ to have a continuous value in the range of $[0,1]$. We will validate this relaxation in our experiments, where we also check the validity of the above assumption of depth estimation from sparse pixels. 

Finally, we formulate our problem as the following optimization:
\begin{equation}
    \min\limits_M \;
    l_{\rm dif} (Y,\hat{Y}) + \lambda 
    \frac{1}{n} \lVert M \rVert_1
    \label{eqn3}
\end{equation}
where $l_{\rm dif}$ is a measure of difference between $Y$ and $\hat{Y}$; $\lambda$ is a control parameter for the sparseness of $M$; $n$ is the number of pixels; and $\lVert M\rVert_1$ is the $\ell_1$ norm (of a vectorized version) of $M$.

\begin{figure}[!t]
\centering  
\begin{tabular}
{p{0.09\textwidth}<{\centering}p{0.09\textwidth}<{\centering}p{0.09\textwidth}<{\centering}p{0.09\textwidth}<{\centering}}
\IncG[ width=0.7in]{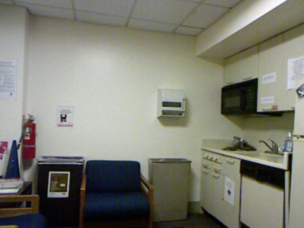}
&\IncG[ width=0.7in]{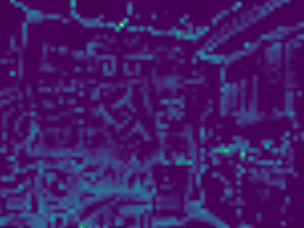}
&\IncG[ width=0.7in]{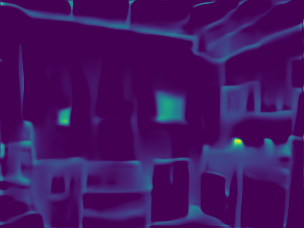}
&\IncG[ width=0.7in]{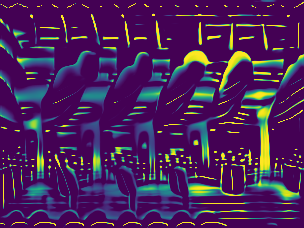}
\\
\IncG[ width=0.7in]{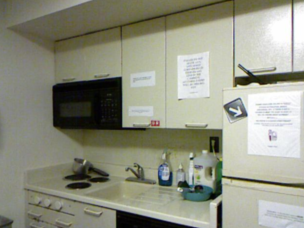}
&\IncG[ width=0.7in]{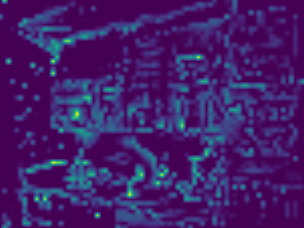}
&\IncG[ width=0.7in]{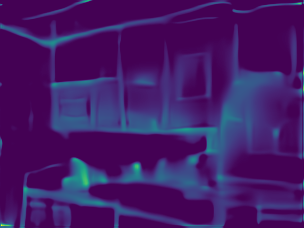}
&\IncG[ width=0.7in]{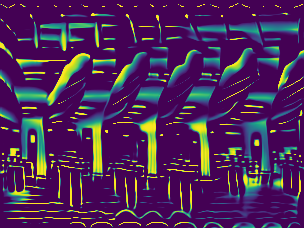}
\\
\IncG[ width=0.7in]{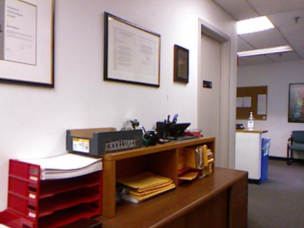}
&\IncG[ width=0.7in]{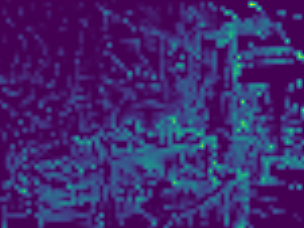}
&\IncG[ width=0.7in]{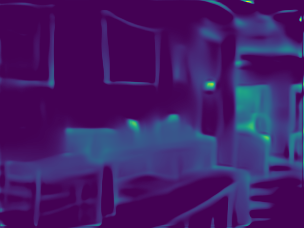}
&\IncG[ width=0.7in]{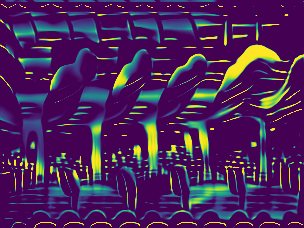}
\\
{\footnotesize (a)} & {\footnotesize (b)} & {\footnotesize (c)} & {\footnotesize (d)}
\end{tabular}
\caption{Form left to right, (a) RGB images (b) $M$ obtained by solving (\ref{eqn3}), (c) $M$ obtained by solving (\ref{eqn4}), (d) $M$ obtained by solving (\ref{eqn5}). }
\label{fig_mask1}
\end{figure}

\subsection{Learning to Predict Mask}
\label{sec_app_fool}

Now we consider how to perform the optimization (\ref{eqn3}). The network $N$ appears in the objective function through the variable $\hat{Y} = N (I \otimes M )$. 
We need carefully consider such optimization associated with the output of a CNN with respect to its input, because it often provides unexpected results, as is shown in previous studies.

In \cite{Simonyan2013DeepInside},
the optimal inputs to CNNs trained on object recognition are computed that maximize the score of a selected object class for the purpose of visualization. Although they provide some insights into what the CNNs have learned, the images thus computed are unstable (\eg., sensitive to initial values); they are distant from natural images and not so easy to interpret.
To obtain more visually interpretable images, researchers have employed several constraints on the input images to be optimized, \eg., the one making them appear to be natural images \cite{googleDeepDream,Nguyen2016MultifacetedFeature}.
In addition, optimization of (a function of) network outputs  sometimes yield unpredictable results; typical examples are the adversarial examples \cite{Fong2017InterpretableEO}.

Thus, instead of minimizing (3) with respect to individual elements of $M$, we use an additional network $G$ to predict $M\approx G(I )$ that minimizes (3). More specifically, we consider the following optimization: 
\begin{equation}
    \min\limits_G \; l_{\rm dif} (Y,N(I \otimes G(I)))+\lambda \frac{1}{n}
    \lVert G(I) \rVert_1,
\label{eqn4}
\end{equation}
where $\lVert G(I)\rVert_1$ indicates $\ell_1$ norm of vectorized $G(I)$. We employ the sigmoid activation function for the output layer of $G$, which constrain its output in the range of $[0,1]$. The details of our method for training $G$ are shown in Algorithm~\ref{alg_leaning_G}.
Figure \ref{fig_mask1} shows comparison of $M$ computed by different methods. It is seen that the direct optimization of (\ref{eqn3}) (Fig.\ref{fig_mask1}(b)) yields noisy, less interpretable maps than our approach (Fig.\ref{fig_mask1}(c)). 

We have considered removing as many unimportant pixels of $I$ as possible while maximally maintaining the original prediction $Y=N(I)$. There is yet another approach to identify important/unimportant pixels, which is to identify the most important pixels of $I$, without which the prediction will maximally deteriorate. This is formulated as the following optimization problem:
\begin{equation}
    \min\limits_G \; -l_{\rm dif} (Y,N(I \otimes G(I)))+\lambda \frac{1}{n}
    \lVert (1 - G(I)) \rVert_1.
\label{eqn5}
\end{equation}
This formulation is similar to that employed in \cite{Fong2017InterpretableEO}, a study for visualization of CNNs for object recognition, in which the most important pixels in the input image are identified by masking the pixels that maximally lower the score of a selected object class. Unlike our method, the authors directly optimize $M$; to avoid artifacts that will emerge in the optimization, they employ additional constraints on $M$ other than its sparseness\footnote{In our experiments, we confirmed that their method works well for VGG networks but behaves unstably for modern CNNs such as ResNets.}. The results obtained by the optimization of (\ref{eqn5}) are shown in Fig.~\ref{fig_mask1}(d). It is seen that this approach cannot provide useful results.

\begin{algorithm}[h]  
  \caption{Algorithm for training the network $G$ for prediction of $M$.}  
  \label{alg_leaning_G}  
  \begin{algorithmic}[1]  
    \Require  
      $N$: a target, fully-trained network for depth estimation;  
      $\psi$: a training set, \ie, pairs of the RGB image and depth map of a scene; 
      $\lambda$: a parameter controlling the sparseness of $M$.
     \renewcommand{\algorithmicrequire}{\textbf{Hyperparameters:}}
    \Require Adam optimizer, learning rate: $1e^{-4}$,  weight decay: $1e^{-4}$, training epochs: $K$.
    \Ensure  
      $G$: a network for predicting $M$.  
     \State Freeze $N$;
    \For{$j$ = 1 to $K$}
        \For{$i$ = 1 to $T$}
            \State Select RGB batch $\psi_i$ from $\psi$;
            \State Set gradients of $G$ to 0;
            \State Calculate depth maps for $\psi_i$:
            \\ \hspace{15ex} $Y_{\psi_i}=N(\psi_i)$;  
            \State Calculate the value (L) of objective function: 
            \\ \hspace{6ex} $L = {l_{\rm dif} (Y_{\psi_i},N(\psi_i \otimes G(\psi_i))}$
            ${+ \lambda \frac{1}{n}\lVert G(\psi_i) \rVert_1}$;  
             \State Backpropagate $L$; 
            \State Update $G$;
        \EndFor
    \EndFor 

  \end{algorithmic}  
\end{algorithm}

\section{Experiments}

\subsection{Experimental Setup}

\paragraph{Datasets} 

We use two datasets NYU-v2 \cite{Silberman2012IndoorSA} and KITTI datasets \cite{Uhrig2017THREEDV} for our analyses, which are the most widely used in the previous studies of monocular depth estimation. The NYU-v2 dataset  contains  464 indoor scenes, for which we use the official splits, 249 scenes for training and 215 scenes for testing.
We obtain approximately 50K unique pairs of an image and its corresponding depth map. Following the previous studies, we use  the same 654 samples for testing. 
The KITTI dataset contains outdoor scenes and is collected by car-mounted cameras and a LIDAR sensor. 
We use the official training/validation splits; there are 86K image pairs for training  and 1K image pairs from the official cropped subsets for testing. 
As the dataset only provides sparse depth maps, we use the depth completion toolbox of the NYU-v2 dataset to interpolate pixels with missing depth.

\begin{figure*}[t]
\centering  
\begin{tabular}
{p{0.088\textwidth}<{\centering}p{0.088\textwidth}<{\centering}p{0.088\textwidth}<{\centering}p{0.088\textwidth}<{\centering}p{0.088\textwidth}<{\centering}p{0.116\textwidth}<{\centering}p{0.116\textwidth}<{\centering}p{0.116\textwidth}<{\centering}}
\IncG[ height=0.53in]{./figures/imgs/out0.png} \vspace*{ -5.6mm}
&\IncG[ height=0.53in]{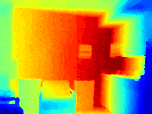} \vspace*{ -5.6mm}
&\IncG[ height=0.53in]{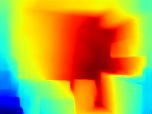} \vspace*{ -5.6mm}
&\IncG[ height=0.53in]{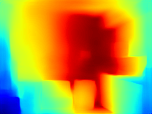} \vspace*{ -5.6mm}
&\IncG[ height=0.53in]{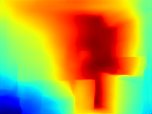} \vspace*{ -5.6mm}
&\IncG[ height=0.72in]{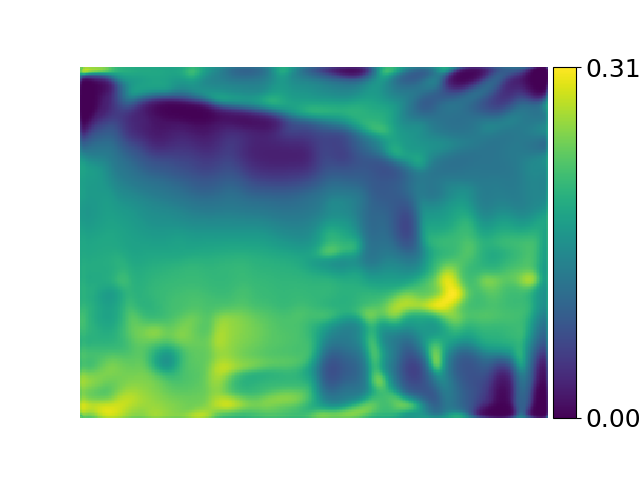} \vspace*{ -5.6mm}
&\IncG[ height=0.72in]{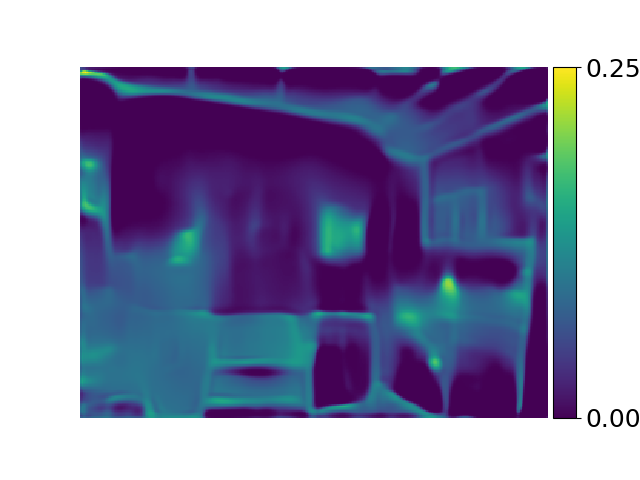} \vspace*{ -5.6mm}
&\IncG[ height=0.72in]{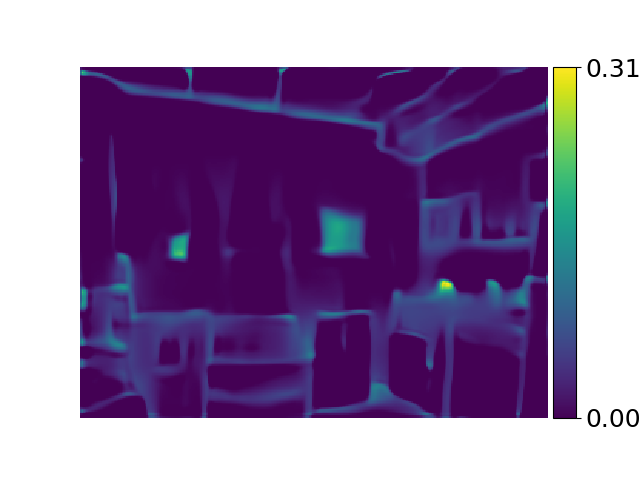} \vspace*{ -5.6mm}
\\
\IncG[ height=0.53in]{./figures/imgs/out1.png} \vspace*{ -5.6mm}
&\IncG[ height=0.53in]{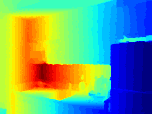} \vspace*{ -5.6mm}
&\IncG[ height=0.53in]{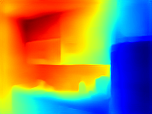} \vspace*{ -5.6mm}
&\IncG[ height=0.53in]{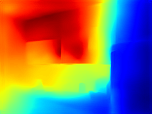} \vspace*{ -5.6mm}
&\IncG[ height=0.53in]{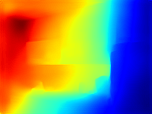} \vspace*{ -5.6mm}
&\IncG[ height=0.72in]{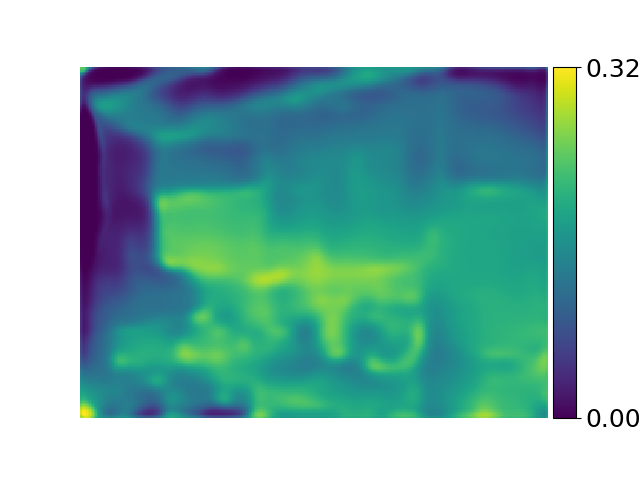} \vspace*{ -5.6mm}
&\IncG[ height=0.72in]{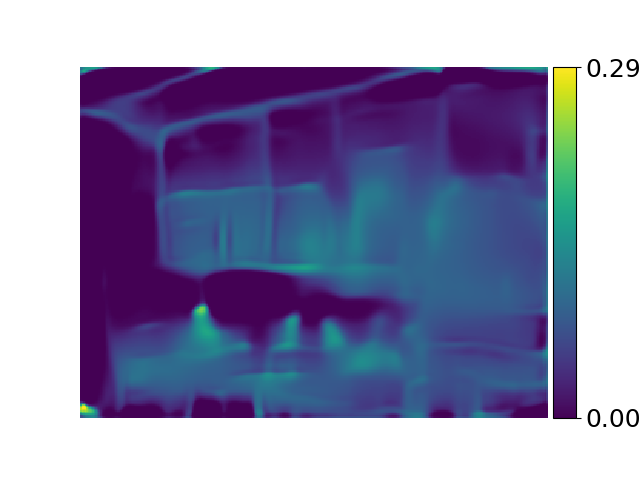} \vspace*{ -5.6mm}
&\IncG[ height=0.72in]{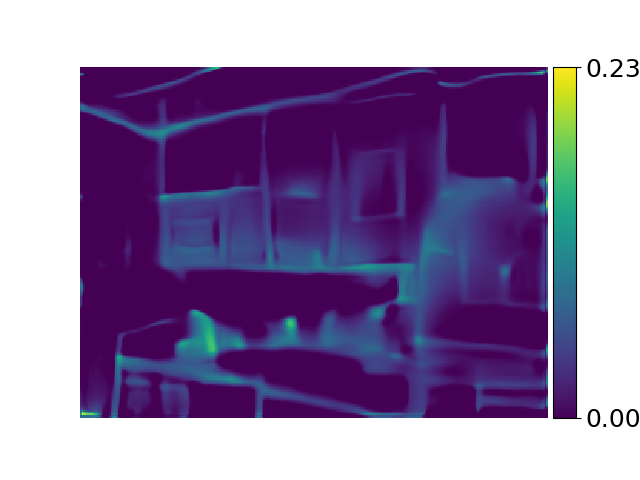} \vspace*{ -5.6mm}
\\
\IncG[ height=0.53in]{./figures/imgs/out2.png} \vspace*{ -5.6mm}
&\IncG[ height=0.53in]{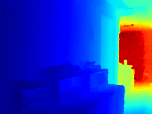} \vspace*{ -5.6mm}
&\IncG[ height=0.53in]{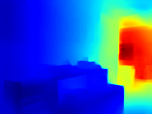} \vspace*{ -5.6mm}
&\IncG[ height=0.53in]{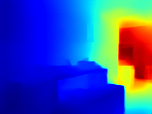} \vspace*{ -5.6mm}
&\IncG[ height=0.53in]{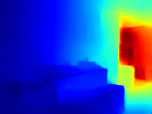} \vspace*{ -5.6mm}
&\IncG[ height=0.72in]{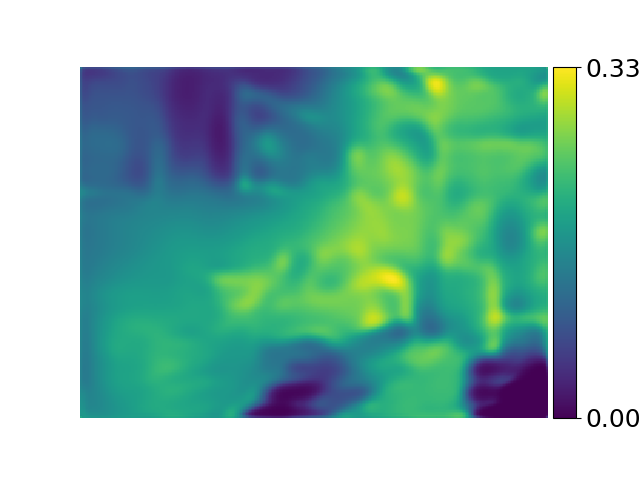} \vspace*{ -5.6mm}
&\IncG[ height=0.72in]{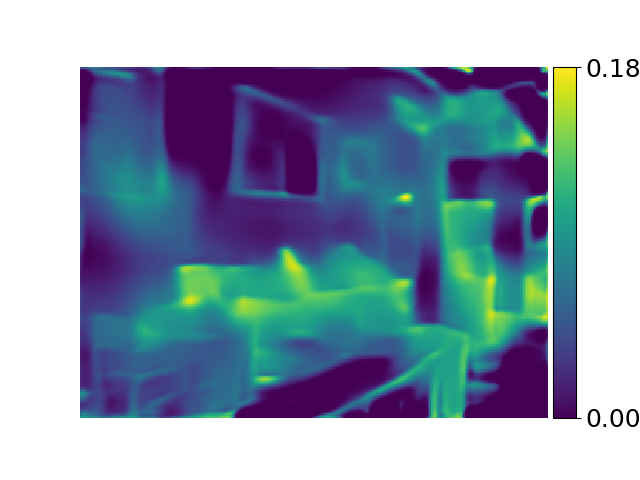} \vspace*{ -5.6mm}
&\IncG[ height=0.72in]{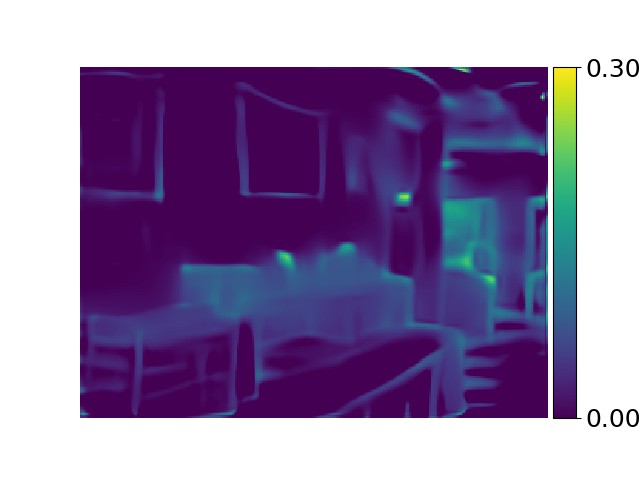} \vspace*{ -5.6mm}
\\
\IncG[ height=0.53in]{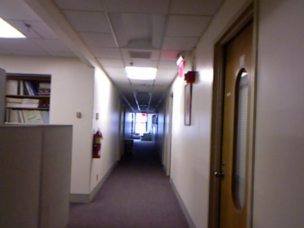} \vspace*{ -4mm}
&\IncG[ height=0.53in]{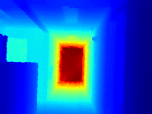}  \vspace*{ -4mm}
&\IncG[ height=0.53in]{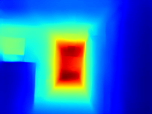} \vspace*{ -4mm}
&\IncG[ height=0.53in]{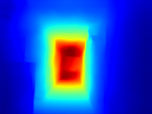} \vspace*{ -4mm}
&\IncG[ height=0.53in]{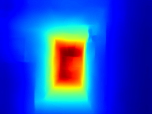} \vspace*{ -4mm}
&\IncG[ height=0.72in]{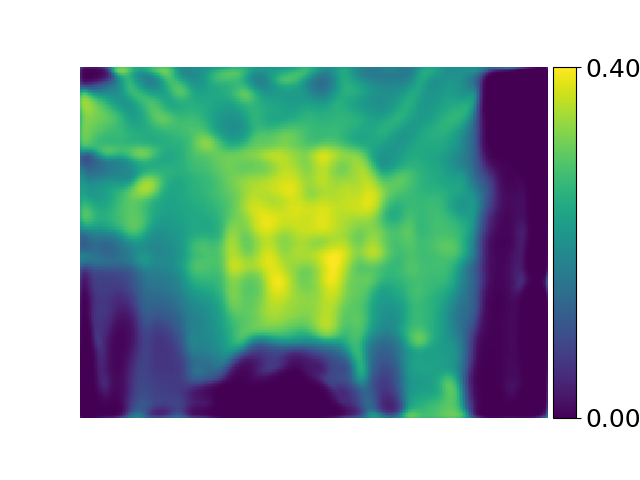} \vspace*{ -4mm}
&\IncG[ height=0.72in]{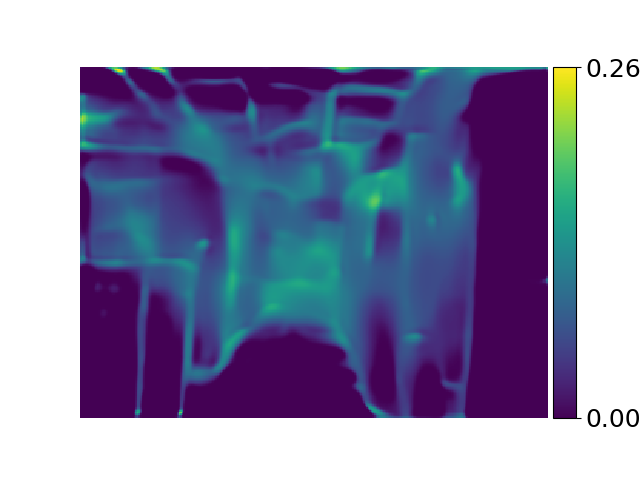} \vspace*{ -4mm}
&\IncG[ height=0.72in]{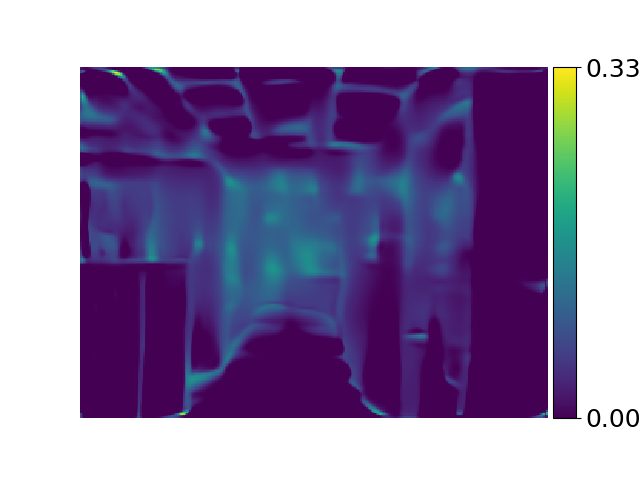} \vspace*{ -4mm}
\\
 {\footnotesize (a) RGB images} & {\footnotesize (b) Ground truth} & {\footnotesize  (c) $\hat{Y}$ when $\lambda$=1}  & {\footnotesize (d) $\hat{Y}$ when $\lambda$=3} &  {\footnotesize (e) $\hat{Y}$ when $\lambda$=5}  & {\footnotesize (f) $M$  when $\lambda$=1} &  {\footnotesize (g) $M$ when $\lambda$=3} &  {\footnotesize (h) $M$ when $\lambda$=5}
\end{tabular}
\vspace*{2mm}
\caption{Visual comparison of approximated depth maps and estimated masks ($M$'s) for different values of the sparseness parameter $\lambda$.}
\label{fig_spasity}
\end{figure*}

\paragraph{Target CNN models}

There are many studies for monocular depth estimation, in which a variety of architectures are proposed. Considering the purpose here, we choose models that show strong performance in estimation accuracy with a simple architecture. 
One is an encoder-decoder network based on ResNet-50  proposed in \cite{laina2016deeper}, which outperforms previous ones by a large margin as of the time of publishing. We also consider more recent ones proposed in \cite{hu2019revisiting}, for which we choose three different backbone networks, ResNet-50 \cite{he2016deep}, DenseNet-161 \cite{huang2016densely}, and SENet-154 \cite{hu2018senet}. 
For better comparison, all the models are implemented in the same experimental conditions. 
Following their original implementation, the first and the latter three models are trained using different losses. To be specific, the first model is trained using $\ell_1$ norm of depth errors\footnote{We have found that $\ell_1$ performs better than the $berhu$ loss originally used  in \cite{laina2016deeper}, which agree with \cite{ma2017sparse}.}.
For the latter three models, sum of three losses are used, \ie, $l_{\rm depth}=\frac{1}{n}\sum_{i=1}^n F(e_i)$, 
$l_{\rm grad}=\frac{1}{n}\sum_{i=1}^n (F(\nabla_{x}(e_i))+F(\nabla_{y}(e_i)))$, and $ l_{\rm normal} = \frac{1}{n}\sum_{i=1}^n
    \left(1-\cos\theta_i
    \right),$ 
where $F(e_i) = \ln(e_i+0.5)$; $e_i = \|y_i - \hat{y_i}\|_1$; $y_i$ and $\hat{y_i}$ are true and estimated depths; and $\theta_i$ is the angle between the surface normals computed from the true and estimated depth map.

\paragraph{Network $G$ for predicting $M$}

We employ an encoder-decoder structure for $G$. For the encoder, we use the dilated residual network (DRN) proposed in \cite{Yu2017}, which preserves local structures of the input image due to a fewer counts of down-sampling. Specifically, we use a DRN with 22 layers (DRN-D-22) pre-trained on ImageNet \cite{imagenet_cvpr09}, from which we remove the last fully connected layer. It yields a feature map with 512 channels and $1/8$ resolution of the input image.
For the decoder, we use a network consisting of three up-projection blocks \cite{laina2016deeper} yielding a feature map with 64 channels and the same size as the input image, followed by 
a $3\times3$ convolutional layer outputting $M$. The encoder and decoder are connected to form the network $G$, which has 25.3M parameters in total. For the loss used to train $G$, we use $l_{\rm dif}=l_{\rm depth} + l_{\rm grad} + l_{\rm normal}$. 

\subsection{Estimating Depth from Sparse Pixels}
\label{sec_sparsity}

\begin{table}[t]
\begin{center}
\caption{Accuracy of depth estimation for different values of the sparseness parameter $\lambda$. Results on the NYU-v2 dataset by the ResNet-50 model of \cite{hu2019revisiting}. Sparseness in the table indicates the average number of non-zero pixels in $M'$.}
\label{tbl:nyu_spa} \small
\begin{tabular}{|l|c|c|c|}
\hline
$\lambda$ & RMSE ($M$) & RMSE ($M'$) & Sparseness \\
\hline
original &0.555 &0.555 &1.0  \\ 
$\lambda=1$ &0.605 &0.568 &0.920 \\ 
$\lambda=2$ &0.668 &0.617 &0.746 \\ 
$\lambda=3$ &0.699 &0.668 &0.589 \\ 
$\lambda=4$ &0.731 &0.733 &0.425  \\ 
$\lambda=5$ &0.740 &0.758 &0.361  \\ 
$\lambda=6$ &0.772 &0.882 &0.215  \\ 
\hline
\end{tabular}
\end{center}
\end{table}

As explained above, our approach is based on the assumption that the network $N$ can accurately estimate depth from only a selected set of sparse pixels. We also relaxed the condition on the binary mask, allowing $M$ to have continuous values in the range of [0,1]. 
To validate the assumption as well as this relaxation, we check how the accuracy of depth estimation will change when binarizing the continuous mask $M$ predicted by $G$. 

To be specific, computing $M=G(I)$ for $I$, we binarize $M$ into a binary map $M'$ using a threshold $\epsilon=0.025$. 
We then compare accuracy of the predicted depth maps $N(I\otimes M')$ and $N(I\otimes M)$. As the sparseness of $M$ is controlled by the parameter $\lambda$ as in Eq.(\ref{eqn4}), we evaluate the accuracy for different $\lambda$'s. We use the NYU-v2 dataset and a ResNet-50 based model of \cite{hu2019revisiting}. We train it for 10 epochs on the training set and measure its accuracy by RMSE.

Table \ref{tbl:nyu_spa} shows the results. It is first observed that there is trade-off between accuracy of depth estimation and sparseness of the mask $M$. Figure \ref{fig_spasity} shows examples of pairs of the mask $M$ and estimated depth map $\hat{Y}$ for different $\lambda$'s for four different input images. It is also observed from Table \ref{tbl:nyu_spa} that the estimated depth with the binarized mask $M'$ is mostly the same as that with the continuous $M$ when $\lambda$ is not too large; it is even more accurate for small $\lambda$'s. This validates our relaxation allowing $M$ to have continuous values. Considering the trade-off between estimation accuracy and $\lambda$ as well as the difference between prediction with $M$ and $M'$, we choose $\lambda=5$ in the analyses shown in what follows.

\subsection{Analyses of Predicted Mask}
\label{sec_vis}

\begin{figure}[t]
\centering
\subfigure {\includegraphics[scale=0.56]{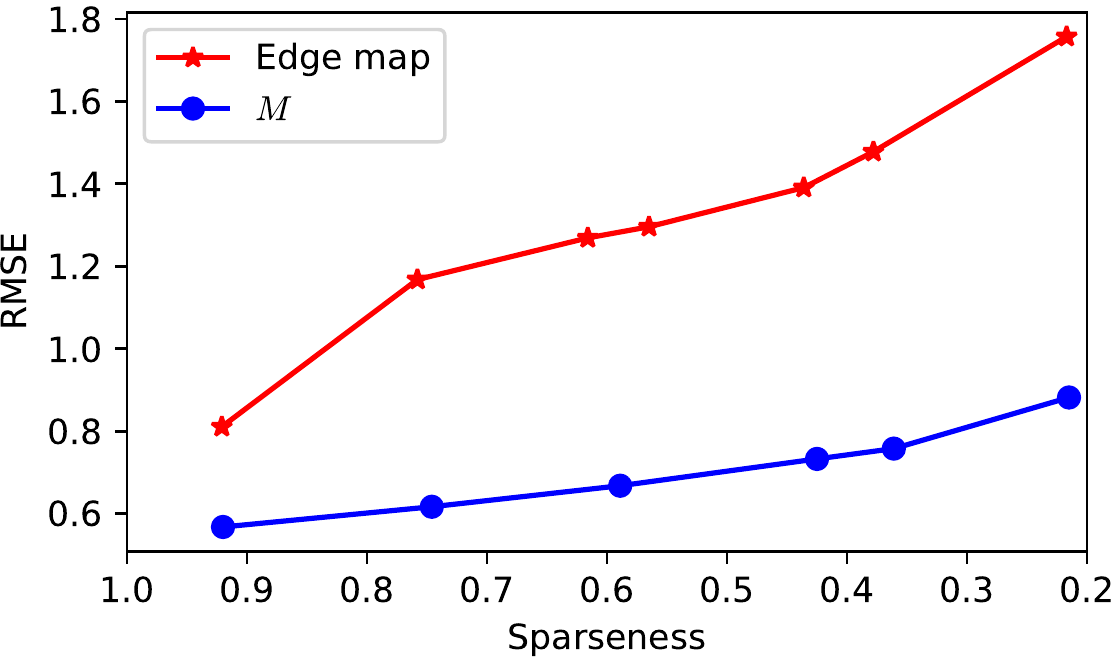}}
\caption{Comparison of accuracy of depth estimation when selecting input image pixels using $M$ and using the edge map of input images.} 
\label{fig_edge_map}
\end{figure}

\begin{figure*}[t]
\centering  
\begin{tabular} 
{p{0.03\textwidth}<{\centering}p{0.1\textwidth}<{\centering}p{0.1\textwidth}<{\centering}p{0.1\textwidth}<{\centering}p{0.1\textwidth}<{\centering}p{0.1\textwidth}<{\centering}p{0.1\textwidth}<{\centering}}
\footnotesize (1)
&\IncG[ width=0.8in]{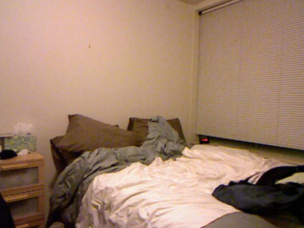}
&\IncG[ width=0.8in]{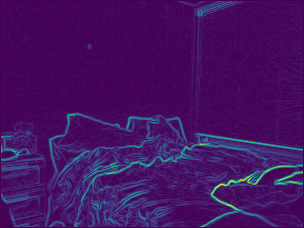}
&\IncG[ width=0.8in]{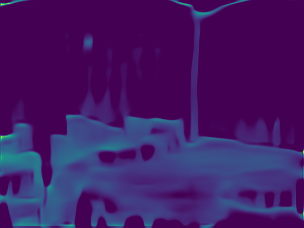}
&\IncG[ width=0.8in]{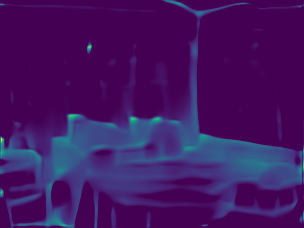}
&\IncG[ width=0.8in]{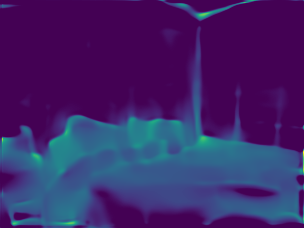}
&\IncG[ width=0.8in]{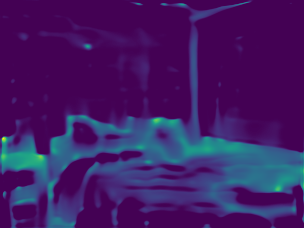} \\
\footnotesize (2)
&\IncG[ width=0.8in]{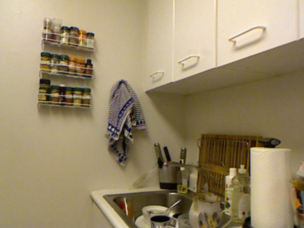}
&\IncG[ width=0.8in]{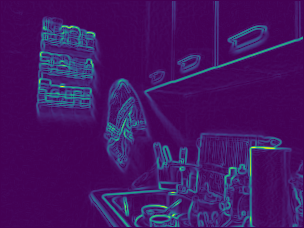}
&\IncG[ width=0.8in]{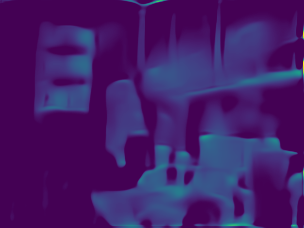}
&\IncG[ width=0.8in]{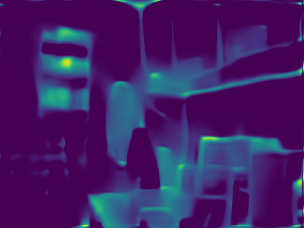}
&\IncG[ width=0.8in]{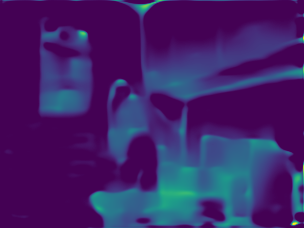}
&\IncG[ width=0.8in]{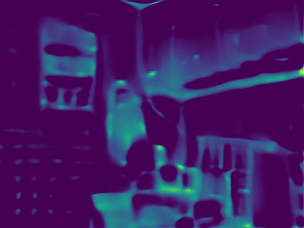} \\
\footnotesize (3)
&\IncG[ width=0.8in]{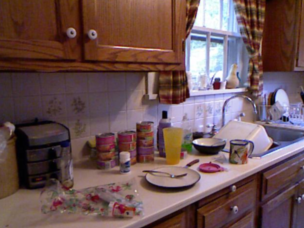}
&\IncG[ width=0.8in]{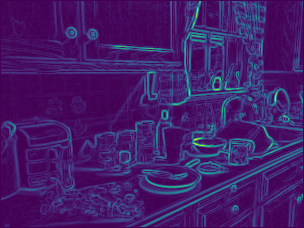}
&\IncG[ width=0.8in]{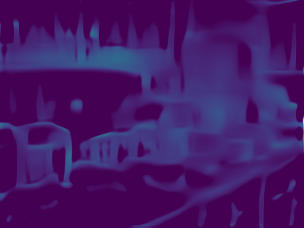}
&\IncG[ width=0.8in]{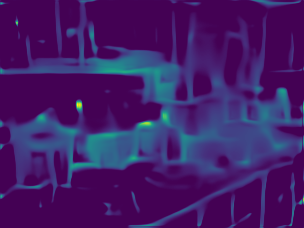}
&\IncG[ width=0.8in]{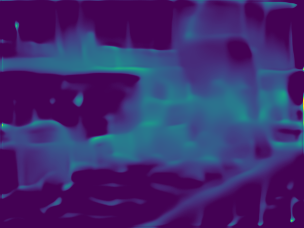}
&\IncG[ width=0.8in]{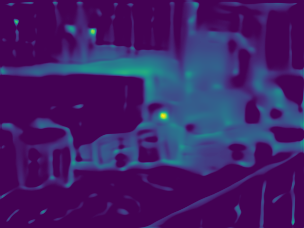} \\
\footnotesize (4)
&\IncG[ width=0.8in]{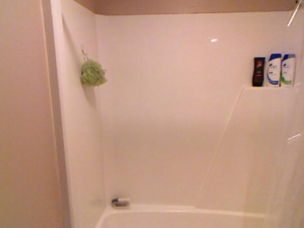}
&\IncG[ width=0.8in]{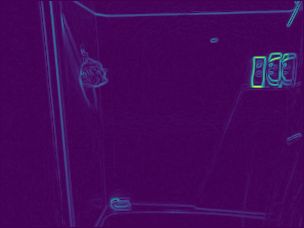}
&\IncG[ width=0.8in]{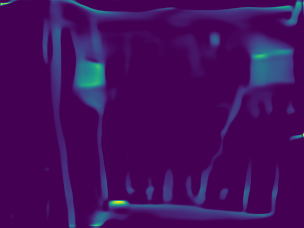}
&\IncG[ width=0.8in]{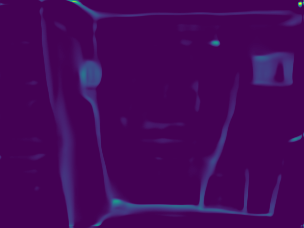}
&\IncG[ width=0.8in]{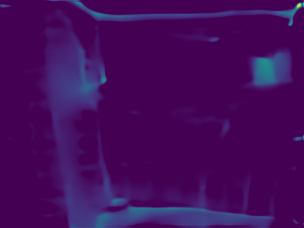}
&\IncG[ width=0.8in]{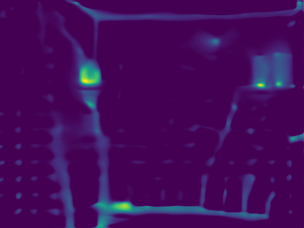} \\
\footnotesize (5)
&\IncG[ width=0.8in]{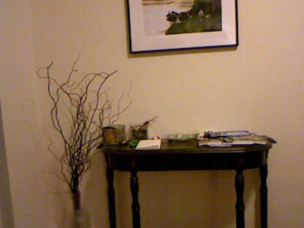}
&\IncG[ width=0.8in]{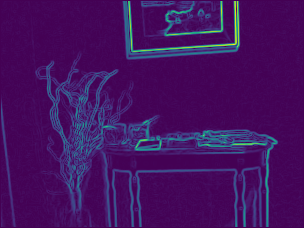}
&\IncG[ width=0.8in]{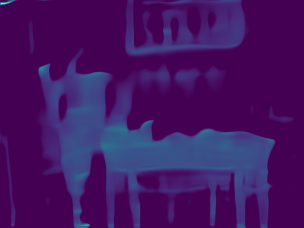}
&\IncG[ width=0.8in]{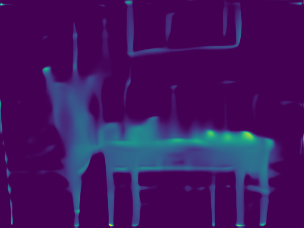}
&\IncG[ width=0.8in]{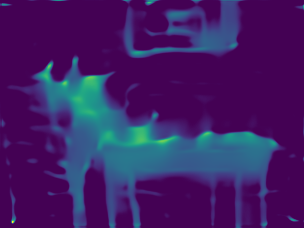}
&\IncG[ width=0.8in]{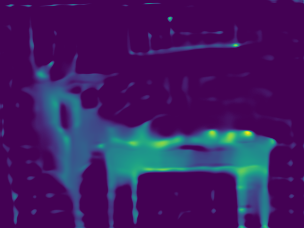} \\
\footnotesize (6)
&\IncG[ width=0.8in]{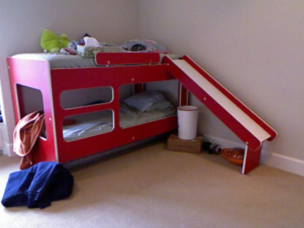}
&\IncG[ width=0.8in]{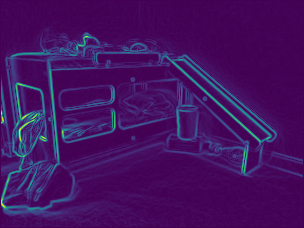}
&\IncG[ width=0.8in]{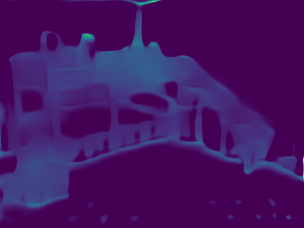}
&\IncG[ width=0.8in]{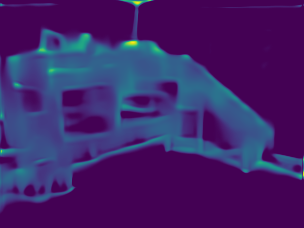}
&\IncG[ width=0.8in]{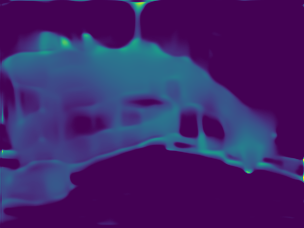}
&\IncG[ width=0.8in]{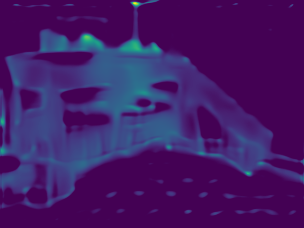} \\
\footnotesize (7)
&\IncG[ width=0.8in]{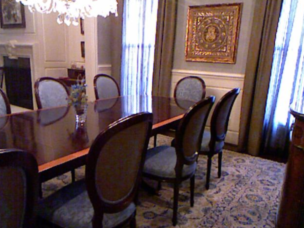}
&\IncG[ width=0.8in]{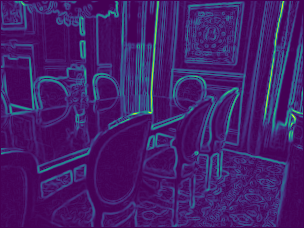}
&\IncG[ width=0.8in]{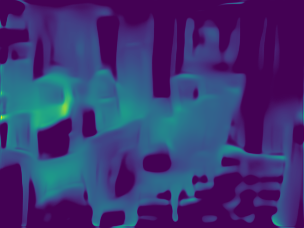}
&\IncG[ width=0.8in]{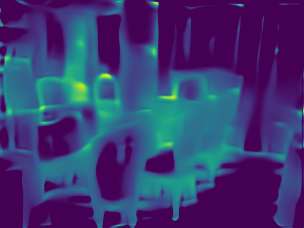}
&\IncG[ width=0.8in]{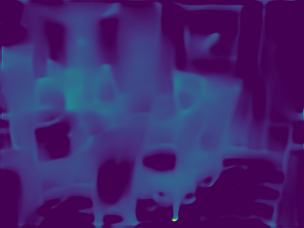}
&\IncG[ width=0.8in]{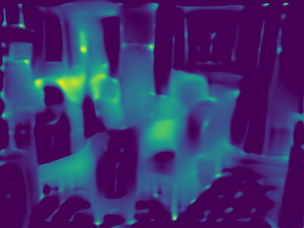} \\
\footnotesize (8)
&\IncG[ width=0.8in]{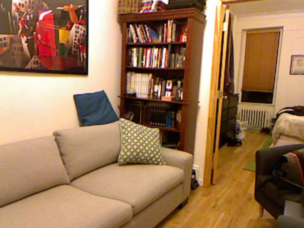}
&\IncG[ width=0.8in]{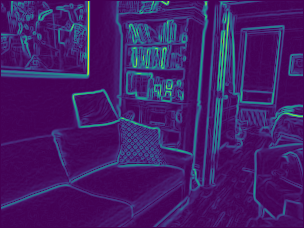}
&\IncG[ width=0.8in]{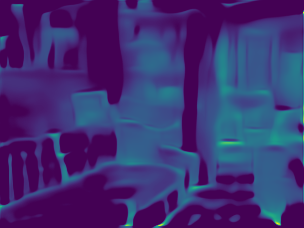}
&\IncG[ width=0.8in]{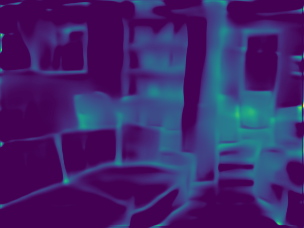}
&\IncG[ width=0.8in]{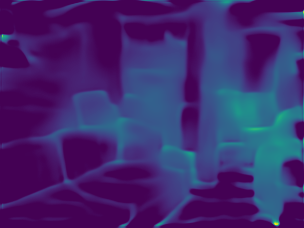}
&\IncG[ width=0.8in]{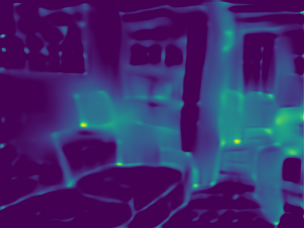} \\



\footnotesize (9)
&\IncG[ width=0.8in]{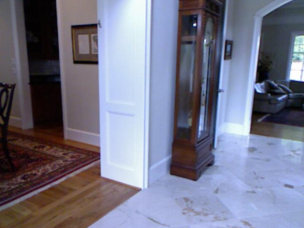}
&\IncG[ width=0.8in]{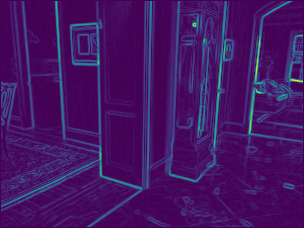}
&\IncG[ width=0.8in]{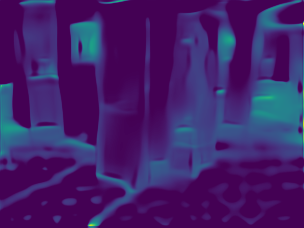}
&\IncG[ width=0.8in]{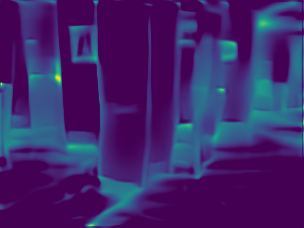}
&\IncG[ width=0.8in]{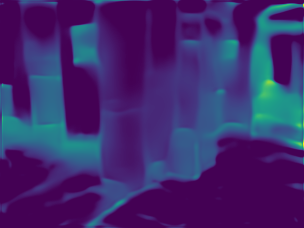}
&\IncG[ width=0.8in]{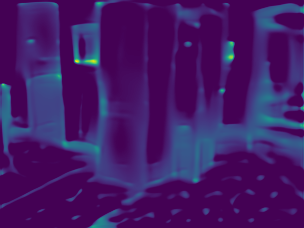} \\
\footnotesize (10)
&\IncG[ width=0.8in]{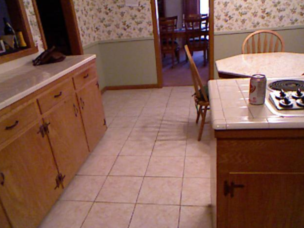}
&\IncG[ width=0.8in]{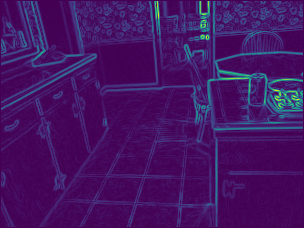}
&\IncG[ width=0.8in]{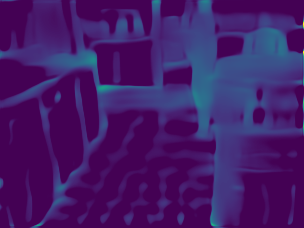}
&\IncG[ width=0.8in]{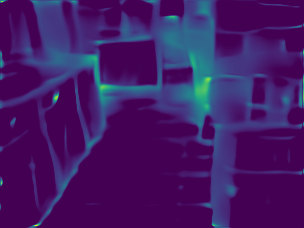}
&\IncG[ width=0.8in]{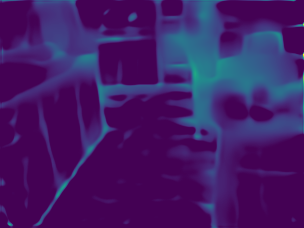}
&\IncG[ width=0.8in]{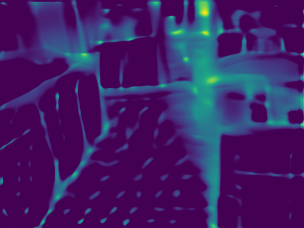} \\
& {\footnotesize (a) RGB images} & {\footnotesize (b) Edge maps}  & {\footnotesize (c) $M$ for \cite{laina2016deeper} (ResNet-50)} &{\footnotesize (d)  $M$ for \cite{hu2019revisiting} (ResNet-50)} &{\footnotesize (e)  $M$ for \cite{hu2019revisiting} (DenseNet-161)} & {\footnotesize (f) $M$ for \cite{hu2019revisiting} (SENet-154)}
\end{tabular}
\vspace*{2mm}
\caption{Predicted masks for different input images for different depth estimation networks, ResNet-50-based model of \cite{laina2016deeper} and three models of \cite{hu2019revisiting} whose backbones are ResNet-50, DenseNet-161, and SENet-154, respectively. The edge map of the input $I$ is also shown for comparison. }
\label{fig_spa_eva_nyu}
\end{figure*}

\begin{figure*}[t]
\centering  
\begin{tabular}
{>{\centering\arraybackslash}m{2cm}>{\centering\arraybackslash}m{4cm}>{\centering\arraybackslash}m{4cm}>{\centering\arraybackslash}m{4cm}}
{\footnotesize RGB images}
&\IncG[  width=1.66in]{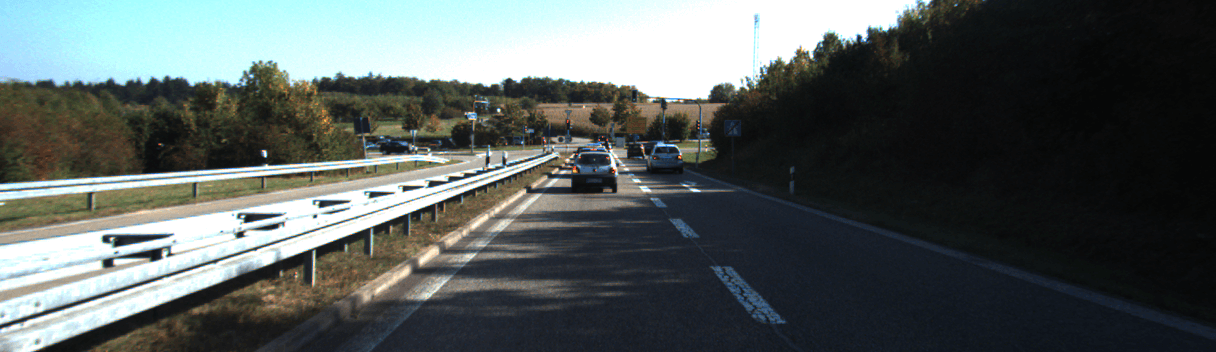}
&\IncG[  width=1.66in]{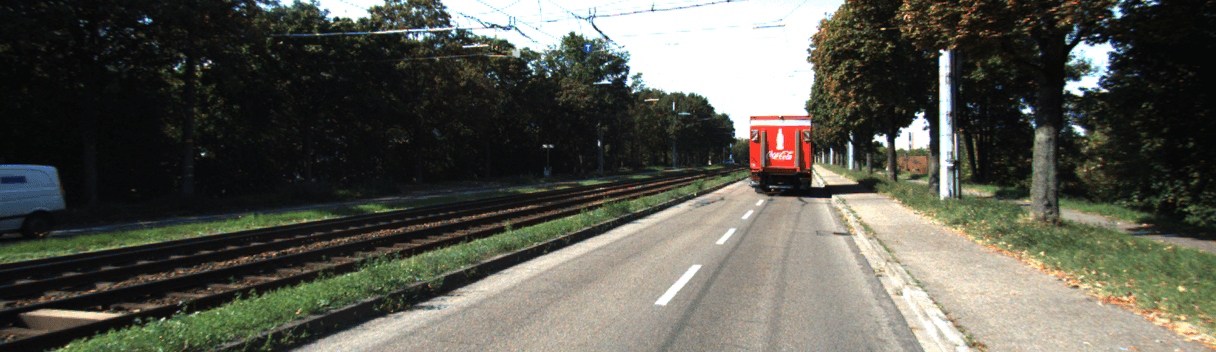}
&\IncG[  width=1.66in]{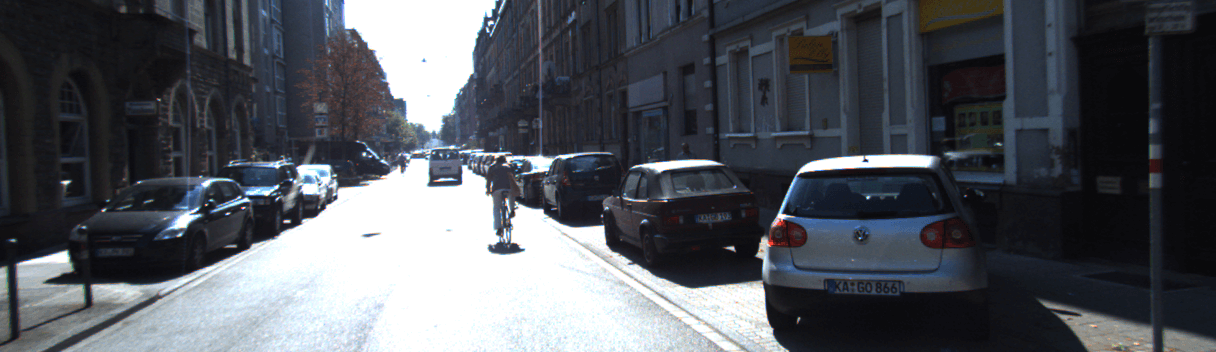}
\\
{\footnotesize Edge maps}
&\IncG[  width=1.66in]{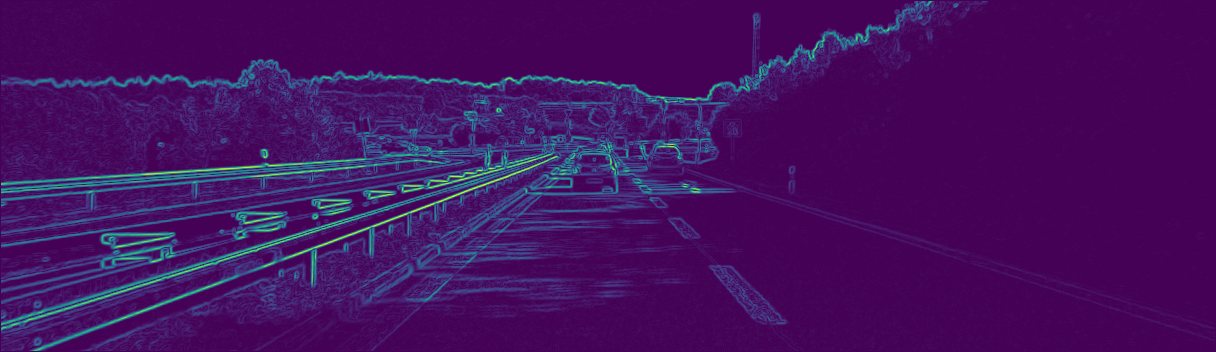}
&\IncG[  width=1.66in]{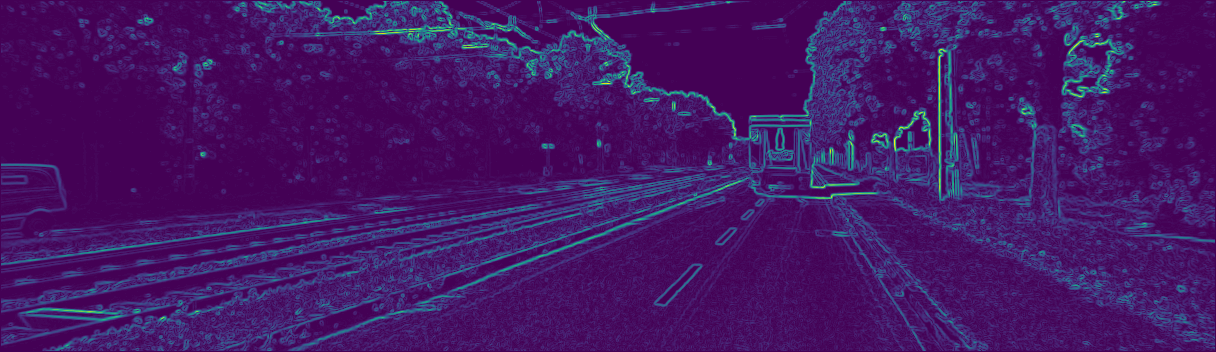}
&\IncG[  width=1.66in]{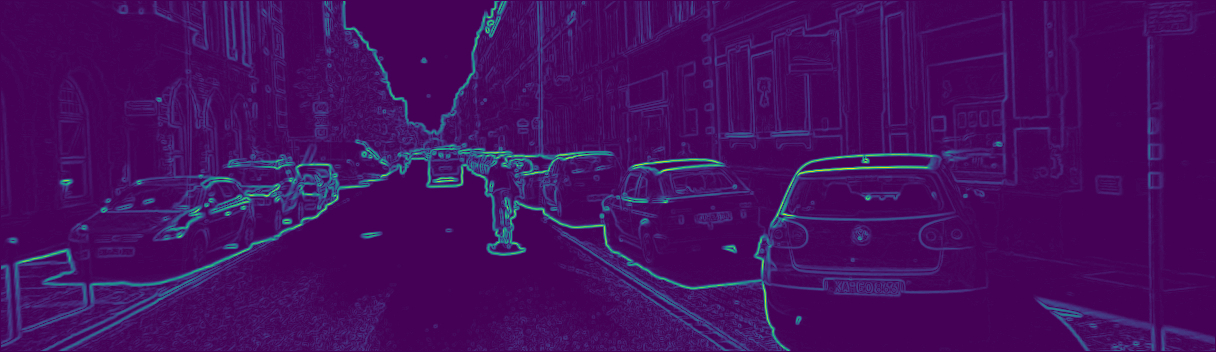}
\\
 {\footnotesize $M$ for \cite{laina2016deeper} (ResNet-50)}
&\IncG[  width=1.66in]{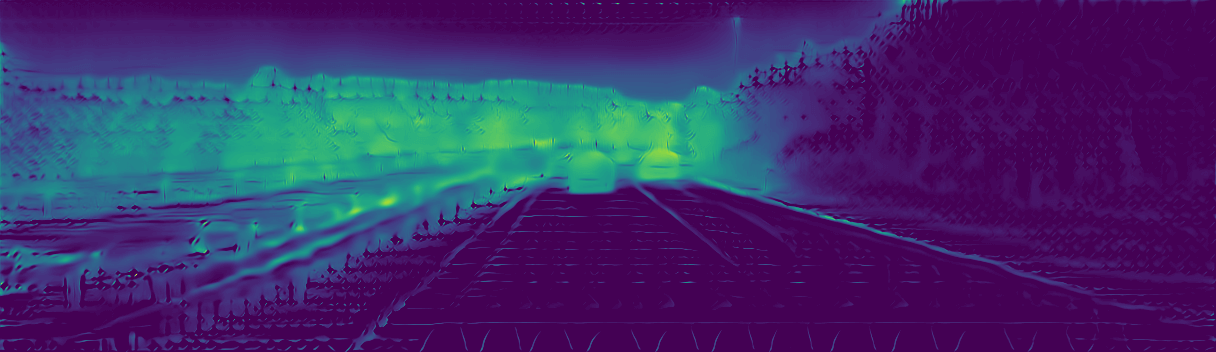} 
&\IncG[  width=1.66in]{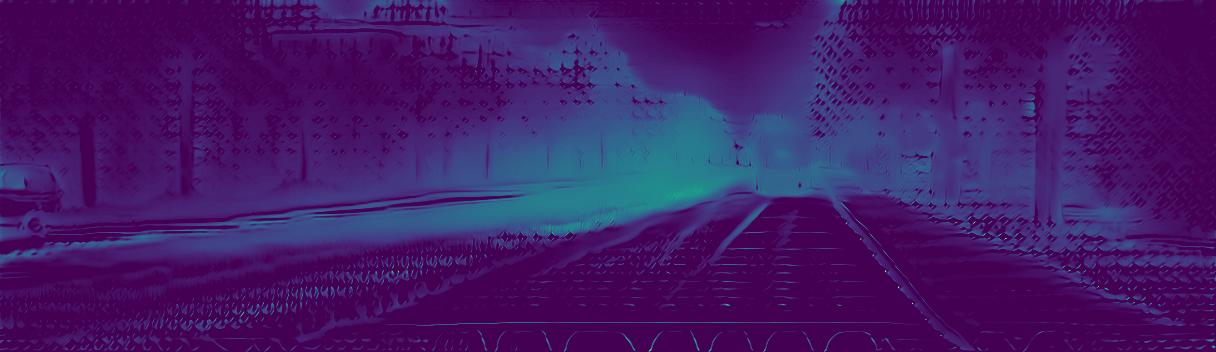}
&\IncG[  width=1.66in]{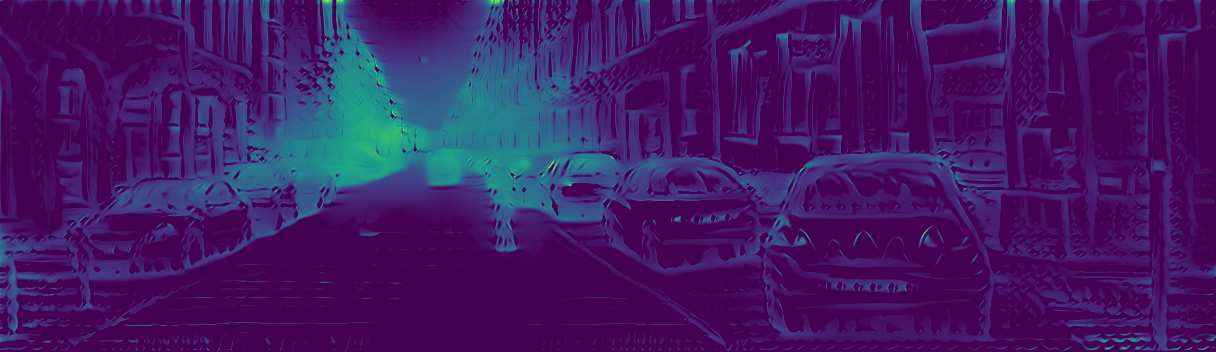} 
\\
 {\footnotesize $M$ for \cite{hu2019revisiting} (ResNet-50)}
&\IncG[  width=1.66in]{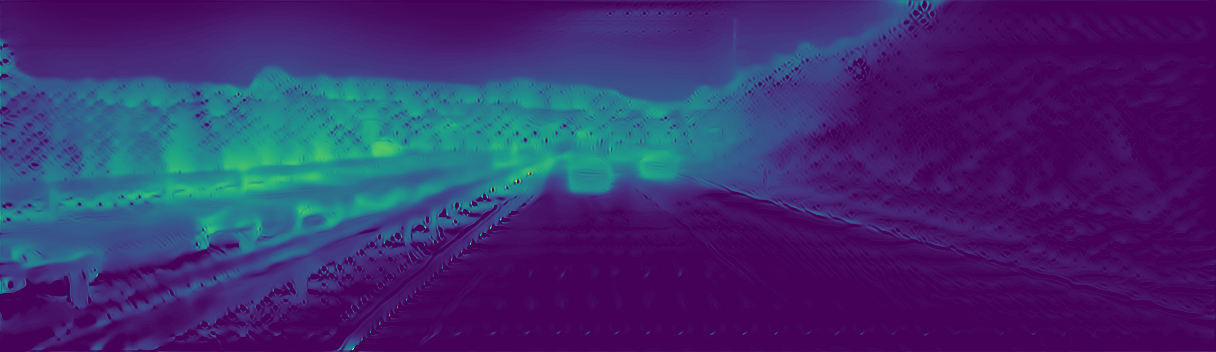} 
&\IncG[  width=1.66in]{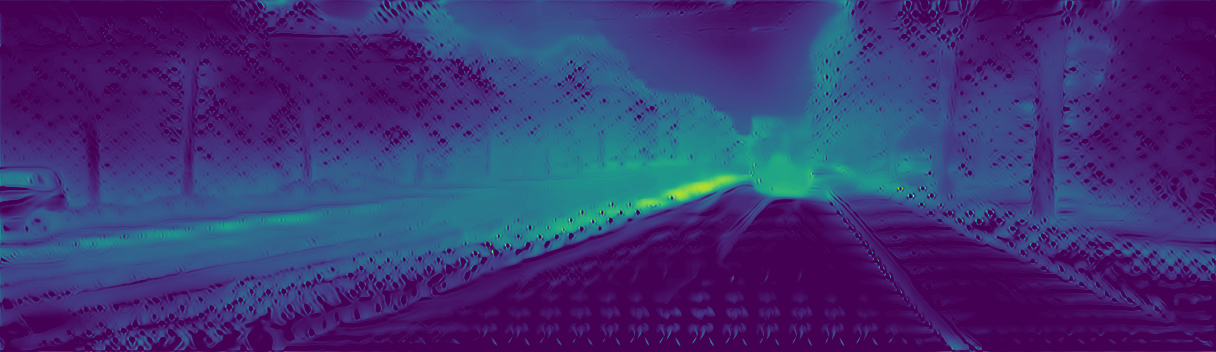}
&\IncG[  width=1.66in]{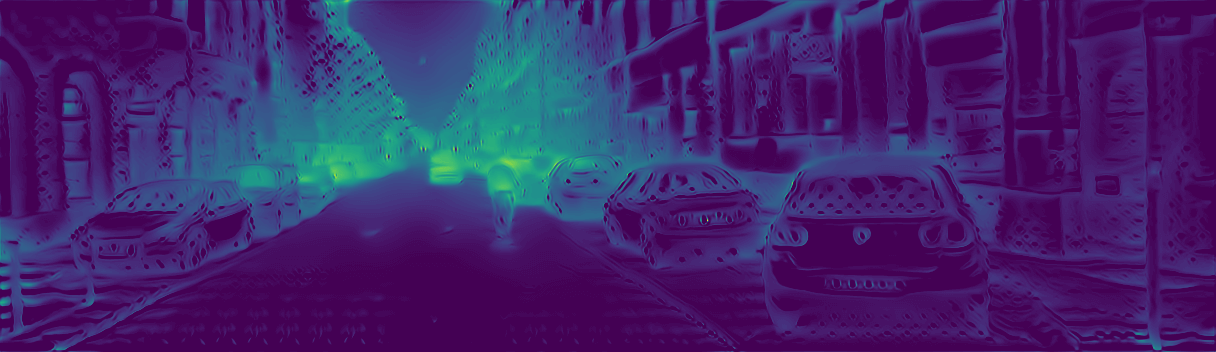} 
\\
 {\footnotesize $M$ for \cite{hu2019revisiting} (DenseNet-161)}
&\IncG[  width=1.66in]{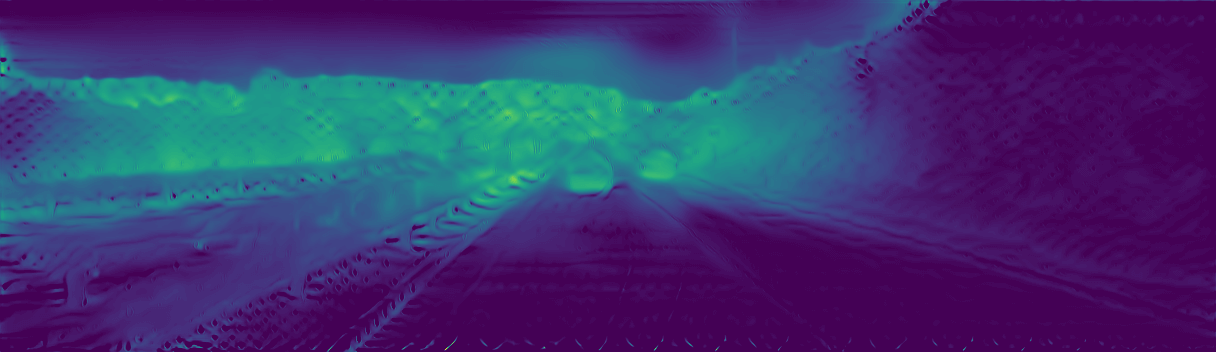}
&\IncG[  width=1.66in]{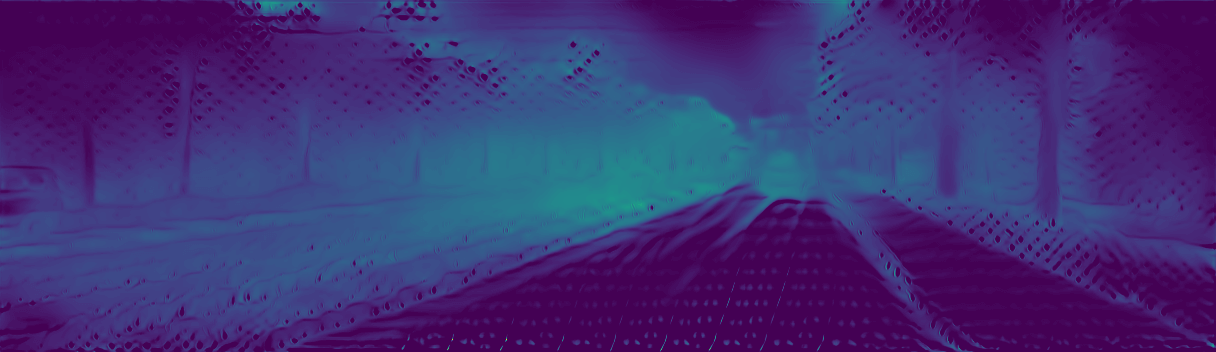}
&\IncG[  width=1.66in]{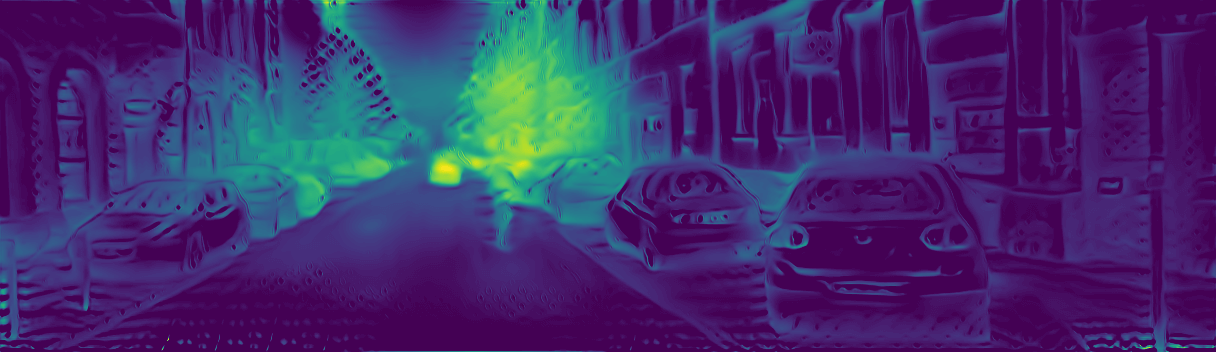}
\\
 {\footnotesize $M$ for \cite{hu2019revisiting} (SENet-154)}
&\IncG[  width=1.66in]{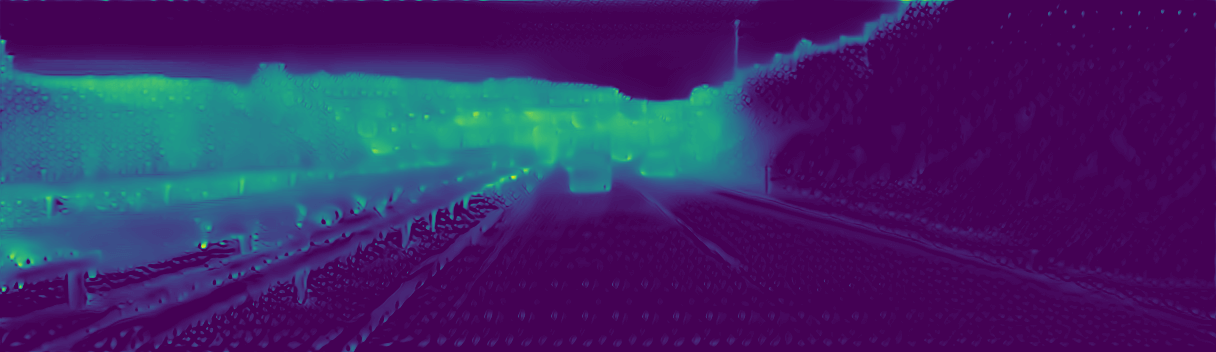}
&\IncG[  width=1.66in]{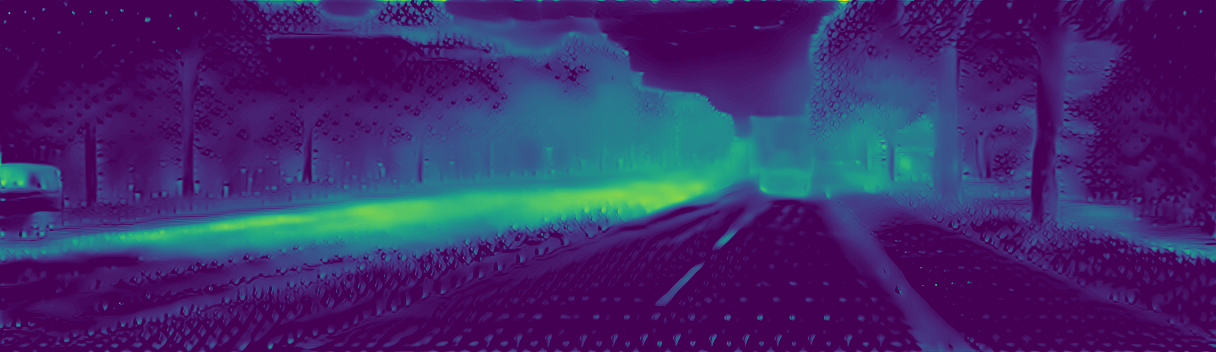}
&\IncG[  width=1.66in]{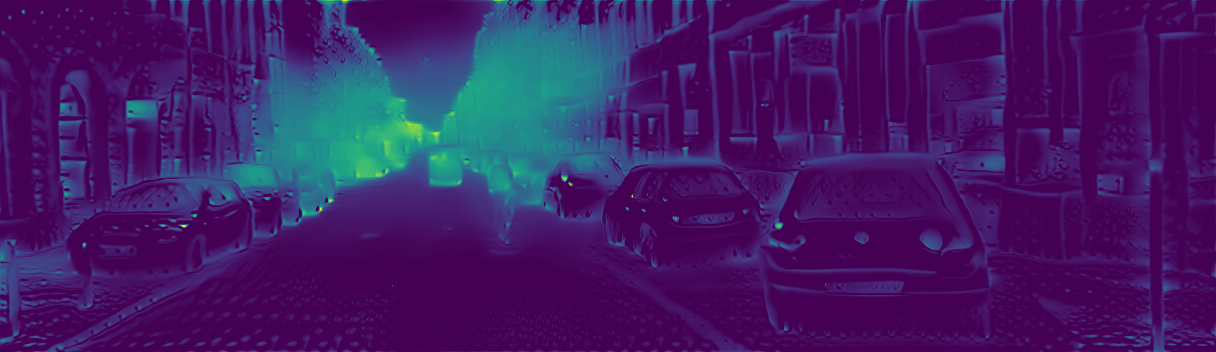}
\\
\end{tabular}
\vspace*{2mm}
\caption{Predicted masks for different networks trained on the KITTI dataset for different input images from the test split. }
\label{fig_spa_eva_kitti}
\end{figure*}

\subsubsection{NYU-v2 dataset}

Figure \ref{fig_spa_eva_nyu} shows predicted masks for different input images and different depth prediction networks. 
It is first observed that there are only small differences among different networks. This will be an evidence that the proposed visualization method can stably identify relevant pixels to depth estimation.
For the sake of comparison, edge maps of $I$ are also shown in Figure \ref{fig_spa_eva_nyu}. 
It is seen from comparison with them that $M$ tends to have non-zero values on the image edges; some non-zero pixels indeed lie exactly on image edges (\eg., the vertical edge on the far side in (1)). 

However, a closer observation reveals that there is also a difference between $M$ and the edge map; 
$M$ tends to have non-zero pixels over the filled regions of objects, not on their boundaries,
as with the table in (5), the chairs in (7) \etc. Moreover, very strong image edges sometimes disappear in $M$, as is the case with a bottom edge of the cabinet in (2); instead, $M$ has non-zero pixels along a weaker image edge emerging on the border of the cabinet and the wall. This is also the case with the intersecting lines between the floor and the bed in (6); $M$ has large values along them, whereas their edge strength is very weak. 

To further investigate (dis)similarity between $M$ and the edge map, we compare them by 
setting the edge map to $M$ and evaluate the accuracy of the predicted depth $N(I\otimes M)$. Figure \ref{fig_edge_map} shows the results. It is seen that the use of edge maps yields less accurate depth estimation, which clearly indicates the difference of the edge maps and the masks predicted by $G$.

Not boundary alone but filled region is highlighted for small objects. We conjecture that the CNNs recognize the objects and somehow utilize it for depth estimation.

\subsubsection{KITTI dataset}

Figure \ref{fig_spa_eva_kitti} shows the predicted masks on the KITTI dataset for three randomly selected images along with their edge maps. More examples are given in the supplementary material. As with the NYU-v2 dataset, the predicted masks tend to consist of edges and filled regions, and are clearly different from the edge maps. It is observed that some image edges are seen in the masks but some are not.
For example, in the first image, the guard rail on the left has strong edges, which are also seen in the mask. On the other hand, the white line on the road surface provides strong edges in the edge map but is absent in the mask. This indicates that the CNNs utilizes the guard rail but does not use the white line for depth estimation for some reason. This is also the same as the white vertical narrow object on the roadside in the second image.

A notable characteristic of the predicted masks on this dataset is that the region around the vanishing point of the scene is strongly highlighted in the predicted masks. This is the case with all the images in the dataset, not limited to the three shown here. Our interpretation of this phenomenon will be given in the discussion below.

\subsubsection{Summary and Discussion}

In summary, there are three findings from the above visualization results. 

\paragraph{Important/unimportant image edges}
Some of the image edges are highlighted in $M$ and some are not. This implies that the depth prediction network $N$ selects important edges that are necessary for depth estimation. The selection seems to be more or less independent of the strength of edges. We conjecture that those selected are essential for inferring the 3D structure (\eg., orientation, perspective \etc.) of a room and a road.

\paragraph{Attending on the regions inside objects}
As for objects in a scene, not only the boundary but the inside region of them tend to be highlighted. This is the case more with smaller objects, although this may be partly attributable to the use of sparseness constraint. Unlike the image edges providing the geometric structure of the scene, we conjecture that the depth estimation network $N$ may `recognize' the objects and use their sizes to infer absolute or relative distance to them. 

\paragraph{Vanishing points}
In the case of outdoor scenes of KITTI, the regions around vanishing points (or simply far-away regions) are always highlighted almost without exception. This shows that these regions are important for $N$ to provide accurate depths. This may be attributable to the fact that distant scene points tend to yield large errors because of the loss evaluating the difference in absolute depths; then such distant scene regions will be given more weights than others. Another possible explanation is that this is due to the natural importance of vanishing points; they are naturally a strong cue to understand geometry of a scene. Although these two explanations appear to be orthogonal, they could be coupled with each other in practice. A possible hypothesis is that CNNs (and/or human vision) learn to look at the vanishing points as they are distant and given more weights. Further investigation will be a direction of future studies.

\begin{figure}[t]
\centering  
\begin{tabular}
{p{0.09\textwidth}<{\centering}p{0.09\textwidth}<{\centering}p{0.09\textwidth}<{\centering}p{0.09\textwidth}<{\centering}}
\IncG[ width=0.7in]{./figures/imgs/out0.png}
&\IncG[ width=0.7in]{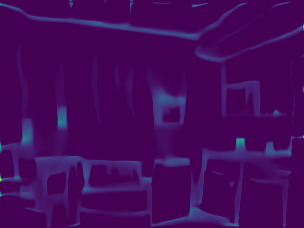}
&\IncG[ width=0.7in]{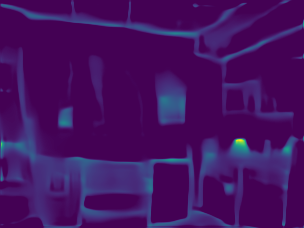}
&\IncG[ width=0.7in]{./figures/res_vis/resnet/mask0.png}
\\
\IncG[ width=0.7in]{./figures/imgs/out1.png}
&\IncG[ width=0.7in]{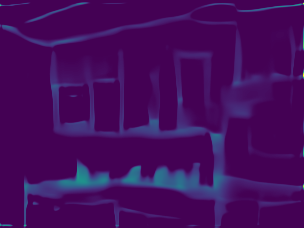}
&\IncG[ width=0.7in]{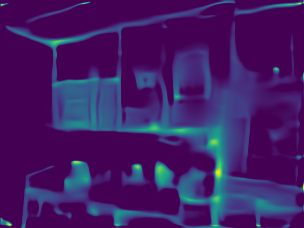}
&\IncG[ width=0.7in]{./figures/res_vis/resnet/mask1.png}
\\
\IncG[ width=0.7in]{./figures/imgs/out2.png}
&\IncG[ width=0.7in]{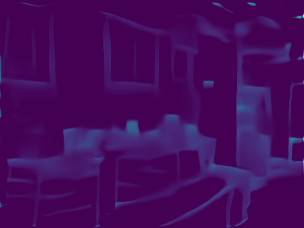}
&\IncG[ width=0.7in]{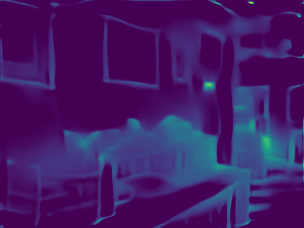}
&\IncG[ width=0.7in]{./figures/res_vis/resnet/mask2.png}
\\
\IncG[ width=0.7in]{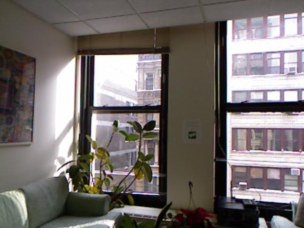}
&\IncG[ width=0.7in]{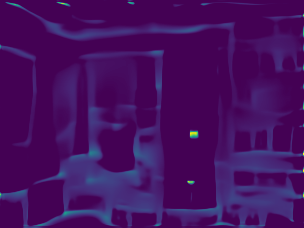}
&\IncG[ width=0.7in]{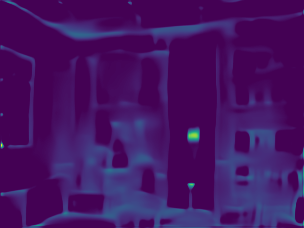}
&\IncG[ width=0.7in]{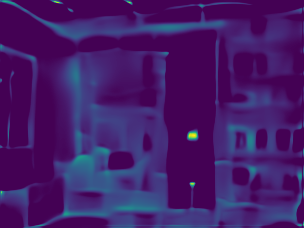}
\\
\IncG[ width=0.7in]{./figures/imgs/out4.png}
&\IncG[ width=0.7in]{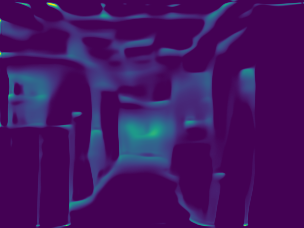}
&\IncG[ width=0.7in]{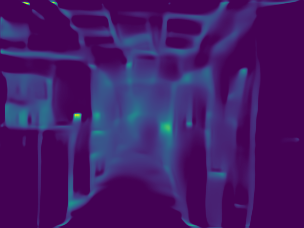}
&\IncG[ width=0.7in]{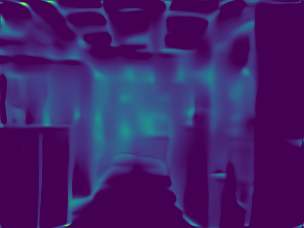}\\
 {\footnotesize (a) RGB images}
&{\footnotesize (b)  $l_{\rm depth}$}
& {\footnotesize (c) $l_{\rm depth}$ $+ l_{\rm grad}$}
& {\footnotesize (d) $l_{\rm depth}$ $+ l_{\rm grad}$ $+ l_{\rm normal}$}
\end{tabular}
\vspace*{2mm}
\caption{Comparison of the estimated mask $M$ for the three combinations of loss functions. }
\label{fig_losses_M}
\end{figure}

\subsection{Evaluation of Training Losses}
\label{sec_loss}

There are several discussions in recent studies on how we should measure accuracy of estimated depth maps \cite{Koch2018EvaluationOC,hu2019revisiting} and what losses we should use for training CNNs \cite{hu2019revisiting}. 
We compare the impact of losses by visualizing a network $N$ trained on different losses. Following \cite{hu2019revisiting}, we consider three losses,  $l_{\rm depth}$ (the most widely used one measuring difference in depth values); $l_{\rm grad}$ (difference in gradients of scene surfaces); and $l_{\rm normal}$ (difference in orientation of normal to scene surfaces). We train a ResNet-50 based model of \cite{hu2019revisiting} on NYU-v2 using different combinations of the three losses, \ie, $l_{\rm depth}$, $l_{\rm depth}+l_{\rm grad}$, and $l_{\rm depth}+l_{\rm grad}+l_{\rm normal}$. 
Figure \ref{fig_losses_M} shows the generated masks for networks trained using the three loss combinations. It is observed that the inclusion of $l_{\rm grad}$ highlights more on the surface of objects. The further addition of $l_{\rm normal}$ highlight more on small objects and makes edges more straight if they should be.


\section{Summary and Conclusion}

Toward answering the question of how CNNs can infer the depth of a scene from its monocular image, we have considered their visualization. Assuming that CNNs can infer a depth map accurately from a small number of image pixels, we considered the problem of identifying these pixels, or equivalently a mask concealing the other pixels, in each input image. We formulated the problem as an optimization problem of selecting the smallest number of pixels from which the CNN can estimate a depth map with the minimum difference to that it estimates from the entire image. Pointing out that there are difficulties with optimization through a deep CNN, we propose to use an additional network to predict the mask for an input image in forward computation. 

We have confirmed through several experiments that the above assumption holds well and the proposed approach can stably predict the mask for each input image with good accuracy. We then applied the proposed method to a number of monocular depth estimation CNNs on indoor and outdoor scene datasets. The results provided several findings, such as i) the behaviour of CNNs that they seem to select edges in input images depending not on their strengths but on importance for inference of scene geometry; ii) the tendency of attending not only on the boundary but the inside region of each individual object; iii) the importance of image regions around the vanishing points for depth estimation on outdoor scenes. We also show an application of the proposed method, which is to visualize the effect of using different losses for training a depth estimation CNN. 

We think these findings contribute to moving forward our understanding of CNNs on the depth estimation task, shedding some light on the problem that has not been explored so far in the community.

{\small
\bibliographystyle{ieee}
\bibliography{egbib}
}

\end{document}